\documentclass[letterpaper, 10 pt, conference]{ieeeconf}

\IEEEoverridecommandlockouts

\usepackage{cite}
\usepackage{amsmath,amssymb,amsfonts}
\usepackage{algorithmic}
\usepackage{graphicx}
\usepackage{textcomp}
\usepackage{xcolor}
\usepackage{url}
\usepackage{cleveref}
\usepackage[caption = false]{subfig}
\usepackage{microtype}
\def\BibTeX{{\rm B\kern-.05em{\sc i\kern-.025em b}\kern-.08em
    T\kern-.1667em\lower.7ex\hbox{E}\kern-.125emX}}
\newcommand\identity{1\kern-0.25em\text{l}}

\definecolor{darkgreen}{rgb}{0.0, 0.6, 0.0}

\begin{document}

\title{\LARGE \bf Improving Visual Place Recognition with\\Sequence-Matching Receptiveness Prediction
}

\author{Somayeh Hussaini\hskip5em Tobias Fischer\hskip5em Michael Milford
\thanks{The authors are with the QUT Centre for Robotics, School of Electrical Engineering and Robotics,  Queensland University of Technology, Brisbane, QLD 4000, Australia. Email: {\tt\footnotesize s.hussaini@qut.edu.au}}
\thanks{This research was partially supported by funding from ARC Laureate Fellowship FL210100156 to MM and ARC DECRA Fellowship DE240100149 to TF. The authors acknowledge continued support from the Queensland University of Technology (QUT) through the Centre for Robotics.}
}

\maketitle

\begin{abstract}

In visual place recognition (VPR), filtering and sequence-based matching approaches can improve performance by integrating temporal information across image sequences, especially in challenging conditions. While these methods are commonly applied, their effects on system behavior can be unpredictable and can actually make performance worse in certain situations. In this work, we present a new supervised learning approach that learns to predict the per-frame sequence matching receptiveness (SMR) of VPR techniques, enabling the system to selectively decide when to trust the output of a sequence matching system. Our approach is agnostic to the underlying VPR technique and effectively predicts SMR, and hence significantly improves VPR performance across a large range of state-of-the-art and classical VPR techniques (namely CosPlace, MixVPR, EigenPlaces, SALAD, AP-GeM, NetVLAD and SAD), and across three benchmark VPR datasets (Nordland, Oxford RobotCar, and SFU-Mountain). We also provide insights into a complementary approach that uses the predictor to replace discarded matches, and present ablation studies including an analysis of the interactions between our SMR predictor and the selected sequence length. 

\end{abstract}

\section{Introduction}

Visual Place Recognition (VPR) and localization are key capabilities for navigating robots and autonomous vehicles~\cite{tsintotas2022revisiting}. Many advances have been made, including in Simultaneous Localization and Mapping~\cite{lajoie2022towards, yan2024gs, zhu2024sni}, Deep Learning~\cite{lecun2015deep}, and most recently Transformers~\cite{han2022survey, vaswani2017attention} and Vision Language Models (VLMs)~\cite{alayrac2022flamingo}. 

One of the key challenges faced by all classes of VPR systems is environmental appearance change, caused by any of a number of factors: lighting conditions, seasons, day-night cycles, and structural change~\cite{masone2021survey, zhang2021visual}. One of the common methods for improving the performance of VPR systems in challenging conditions is to filter input over time and space, for example, through sequence matching techniques~\cite{milford2012seqslam, schubert2021fast, naseer2014robust, garg2022seqmatchnet, mereu2022learning, garg2021seqnet, arroyo2015towards}. But sequence matching is not a perfect solution: sometimes using sequence matching can degrade performance, depending on factors like varying platform velocities or path divergence~\cite{schubert2023visual}. Knowing when to use and trust sequence matching as a VPR performance enhancement technique is the focus of the research presented in this paper.

\begin{figure}[t]
\centering
\vspace*{0.18cm}
\includegraphics[width=0.95\linewidth,trim={0mm 0mm 0mm 0mm},clip]{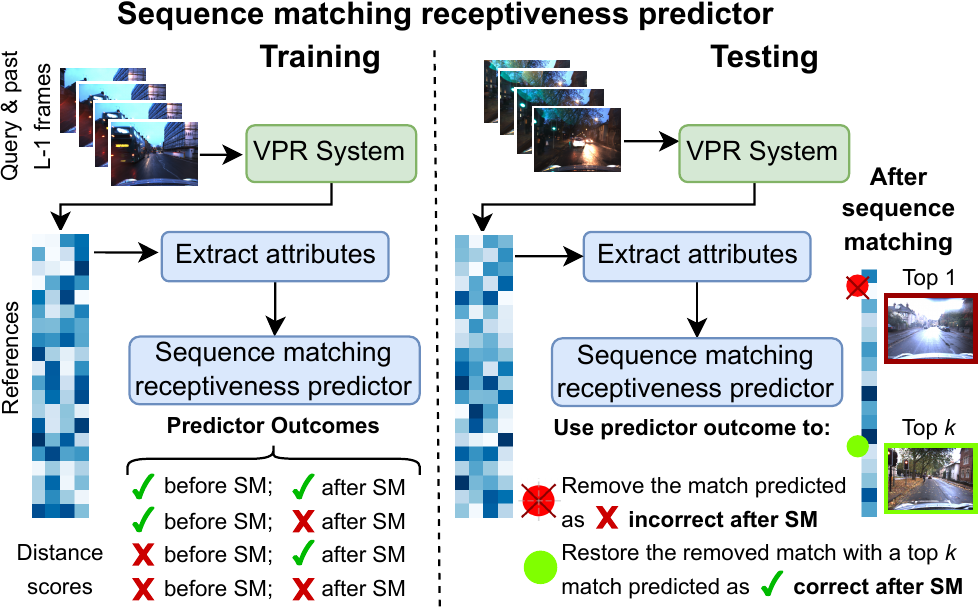} %
\vspace*{-0.2cm}
\caption{Our sequence matching receptiveness predictor uses attributes extracted from the distance scores of a query image and its past $L-1$ frames, where $L$ is the sequence length, to predict the correctness of the sequence matching outcome. Our predictor, applied to a VPR system with sequence matching, removes incorrect matches.}
\vspace*{-0.5cm}
\label{fig:coverfig}
\end{figure}

We present techniques for training a system to predict whether the result of applying sequence matching improves the likelihood of the match being correct, and use this predictor to selectively decide whether to trust the output of a sequence matcher (\Cref{fig:coverfig}). 
As our method is focused on assessing the performance contribution of the sequence matching specifically, it is agnostic to the underlying VPR technique. This general applicability is demonstrated in extensive experimental evaluations, where we evaluate our system across seven VPR techniques, including both recent state-of-the-art methods and classical approaches, and three benchmark datasets. These experiments demonstrate that our predictor, even though not perfect, is able to substantially improve average VPR performance across this range of dataset and method combinations. %

Ablations and analysis also reveal insights into the effect of varying the sequence length when this predictor is applied, and show that the converse to our main contribution – after match rejection, restoring lower-ranked matches based on the predictor – is a much more challenging problem.

Whilst introspection for VPR is a fledgling but growing field, all prior work~\cite{carson2022predicting, claxton2024improving, carson2023unsupervised} has primarily focused on assessment at the instantaneous match level: our contribution here is the first to apply to the widely relevant scenario where filtering or sequence matching is applied.

The paper proceeds as follows. Section II reviews related work in visual place recognition, sequence matching, and introspection. 
Section III details our methodology, including processing the VPR system’s output to train and apply a match quality predictor, and using predictions to remove (and potentially restore) VPR matches. Section IV covers the experimental setup across seven methods and three benchmark datasets. Section V presents overall performance, ablations, and analysis. Section VI concludes with discussion and future work.
Our code is available online\footnote{\url{https://github.com/QVPR/vpr-smr-predictor}}.

\section{Related works}

This section covers related works in visual place recognition (VPR), sequence matching, and introspection for VPR.

\subsection{Visual Place Recognition}

Many VPR approaches use an image retrieval approach which involves extracting meaningful feature representations from both reference and query images~\cite{masone2021survey}. The goal is for the true reference image to be the match with the highest similarity for a given query image, despite challenges caused by appearance, viewpoint, and perceptual aliasing~\cite{schubert2023visual}.

Early VPR methods relied on hand-crafted feature representations~\cite{jegou2010aggregating}, which later evolved into deep learning representations to improve feature robustness~\cite{Arandjelovic2018}. 
More recent VPR methods demonstrate even greater robustness to challenging conditions~\cite{berton2022rethinking, berton2023eigenplaces, ali2023mixvpr, izquierdo2024optimal, lu2024cricavpr, keetha2023anyloc}, which is enabled by the use of Transformer-based architectures, pre-trained or fine-tuned foundation models~\cite{keetha2023anyloc, izquierdo2024optimal, lu2024cricavpr}, feature fusion techniques~\cite{ali2023mixvpr}, and training over multiple image groups~\cite{berton2022rethinking, berton2023eigenplaces}. 
However, these VPR methods can still be susceptible to sudden environmental changes, the appearance of dynamic objects, and perceptual aliasing, which can lead to incorrect matches.

\subsection{Sequence Matching for Visual Place Recognition}

Sequence matching and filtering are common strategies that are widely applied to improve the robustness of single-frame VPR methods. These techniques use the existing temporal information in scenarios where VPR is typically applied, e.g. when a mobile robot moves along a path~\cite{schubert2023visual}. 

Similarity-based sequence matching techniques are applied as a post-processing step to the similarity or distance scores obtained from the matching process~\cite{schubert2023visual, milford2012seqslam, schubert2021fast, naseer2014robust} such as \mbox{SeqSLAM}~\cite{milford2012seqslam}. 
These approaches aggregate matching scores from reference and query subsets, defined by a predefined sequence length~\cite{schubert2023visual}.
Feature-based techniques aggregate consecutive reference and query image features into a single feature representing the entire sequence~\cite{garg2021seqnet, arroyo2015towards}.
Learning-based approaches train sequential feature descriptors to capture temporal information, such as in SeqMatchNet~\cite{garg2022seqmatchnet} and deep-learning or transformer-based methods~\cite{mereu2022learning}.

These sequence matching approaches take advantage of temporal data, which can potentially be more compute-intensive and have higher latency compared to single-frame VPR techniques. Most importantly to the context of this paper, while these methods often improve overall performance, there is typically no self-characterization of whether the sequence matcher improves performance on a match-by-match basis.

\subsection{Introspection for Visual Place Recognition}
\label{LR:introspect}

When deploying VPR systems in real-world environments, a useful capability is the ability for the system to introspect its performance and provide a confidence level in its best match outcome.
The term used to describe the trustworthiness of a localization system's output within an acceptable error tolerance is localization integrity~\cite{al2020localization}, which is important for VPR applications in safety-critical scenarios such as autonomous vehicles~\cite{zhu2022integrity}. 
Localization integrity can be implemented in two ways: reactively, identifying errors after they occur~\cite{al2020localization, gautam2022method, arana2020localization}, or proactively, predicting the outcome of an operation beforehand~\cite{guruau2018learn}. 

Localization integrity has been explored in GNSS navigation systems~\cite{zhu2022integrity}, and introspection of autonomous systems~\cite{gautam2022method, arana2020localization}.
Recently, in the context of VPR, researchers have explored probabilistic approaches to estimate uncertainty and predict the integrity of single-frame systems, both on benchmark datasets~\cite{carson2022predicting} and in real-time robotics~\cite{claxton2024improving}, to improve matching outcomes.
Another study~\cite{carson2023unsupervised} predicted matches within ground-truth tolerance and weighted the scores of likely true single-frame matches to make the VPR system more responsive to sequence matching.

Similar to~\cite{carson2022predicting, claxton2024improving, carson2023unsupervised}, we consider predicting the likelihood that the best match is a true match. While those works focus on predicting or improving single-frame VPR outcomes, our work predicts, on a per-frame basis, whether a single-frame VPR technique would benefit from sequence matching. 
Unlike most SLAM methods, which proactively fuse sensory and motion information over time, our method retrospectively assesses the likely quality of the proposed match after sequence integration has already occurred.

\section{Methodology}

Our approach focuses on predicting sequence matching receptiveness (SMR) on a per-frame basis, and is agnostic to the underlying VPR model. %
The goal is to predict the impact of applying sequence matching to a query prediction: would the prediction remain unchanged; would an incorrect prediction be corrected; or, finally, would a correct prediction turn into an incorrect one?

We frame this as a classification task with a one-to-one correspondence between reference and query images. Our approach predicts sequence matching receptiveness by extracting attributes from the distance scores of a query and its previous frames (with their number defined by the sequence length) and feeding them into a supervised Multi-layer Perceptron (MLP) classifier to predict the query's outcome after sequence matching. We refer to features for our SMR predictor as attributes to distinguish them from image features extracted by VPR models.

\subsection{Distance Matrix}

Here, we detail how the distance matrices from single-frame VPR techniques are constructed, which are used for extracting classifier attributes in~\Cref{M:attributes}.
Given a set of $R$ reference and $Q$ query images, we construct an $R \times Q$ distance \mbox{matrix $D$}, where each element $D(i,j)$ represents the distance scores between the image feature representations of the $i$-th reference image and the $j$-th query image. Common algorithms used for calculating these distances include Euclidean and Cosine distances. 
Our SMR prediction model relies on the distance scores of a query frame and its previous frames to all reference images obtained from the single-frame VPR technique. This is essentially a vertical slice of a distance matrix, and the number of previous frames is determined by the selected sequence matching length $L$.

\subsection{VPR with Sequence Matching}

We incorporate sequence matching into the VPR models using the convolution-based SeqSLAM~\cite{milford2012seqslam}, as presented in SeqMatchNet~\cite{garg2022seqmatchnet}. Here, we provide the necessary background on its formulation for our SMR approach.

Given the single-frame distance matrix, $D$, we apply sequence matching to obtain the sequence distance matrix, $D_\text{seq}$. 
Specifically, the sequence distance score between the reference frame at row $i$ and the query frame at column $j$ is computed as follows: 

\vspace*{-0.35cm}
\begin{equation}
    D_\text{seq}(i, j) = \sum_{y=0}^{L-1} \sum_{x=0}^{L-1} D(i-y, j-x) \identity(y,x),
\end{equation}
where $\identity$ is an $L \times L$ identity-based kernel, and $L$ is the sequence length. 
This approach assumes one-to-one temporal alignment between reference and query frames, as the convolution of an identity kernel with the single-frame distance matrix enforces a linear temporal matching process.

\subsection{SMR Predictor Attributes}
\label{M:attributes}

This section describes the per-frame attributes extracted from the distance scores of a VPR method. 
We define $D_L^j \in \mathbb{R}^{R \times L}$ as the distance scores matrix, which is a slice of $D$ containing scores for the current query frame $j$ and its previous $L-1$ frames. 
We calculate the attributes based on the diagonal entries matrix, $D_\text{Diag}^j \in \mathbb{R}^{(R - L + 1) \times L}$. These entries, spanning from the top-left to bottom-right of the distance scores matrix $D_L^j$, capture sequential correspondences between the query and reference images over the sequence length. 
For each row of diagonal entries $i$, $D_\text{Diag}^j(i, :)$ of query $j$ is defined as:

\vspace*{-0.2cm}
{\small
\begin{equation}
    D_\text{Diag}^j(i, :) = { D_L^j(i, i), D_L^j(i+1, i+1), \dots, D_L^j(i+L, i+L) }.
\end{equation}
}
\vspace*{-0.3cm}

\begin{figure}[t]
\centering
\vspace*{0.22cm}
\includegraphics[width=0.95\linewidth,trim={0mm 0mm 0mm 0mm},clip]{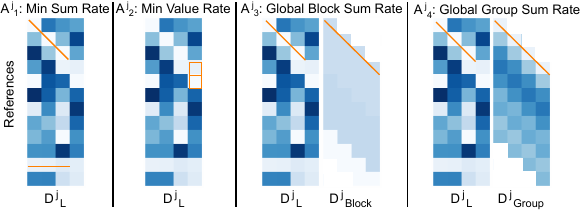} %
\vspace*{-0.3cm}
\caption{
Visualization of the distance scores matrix ($D_{L}^j$) with its corresponding diagonal block entries ($D_\text{block}^j$) and diagonal group entries ($D_\text{group}^j$). Orange lines and squares show the entries used for computing each attribute.}
\vspace*{-0.5cm}
\label{fig:classifier_attributes}
\end{figure}

Extracted attributes from the distance scores matrix for a frame are (\Cref{fig:classifier_attributes}):

\subsubsection{Minimum Sum Rate}
To capture sequential alignment consistency between the query's previous $L-1$ frames, we compute the ratio between the sum of horizontal entries (with the top $k$ lowest row-wise mean) to the sum of diagonal entries (with the top $k$ lowest diagonal-wise mean) from the distance scores matrix, $D_L^j$. 
For horizontal entries in row $i$ of query $j$, the mean distance score across the $L-1$ previous frames is computed as:

\vspace*{-0.4cm}
\begin{equation}
    \mu_\text{Hor}^j(i) = \frac{1}{(L-1)} \sum_{\ell=1}^{(L-1)} D_L^j(i, \ell).
\end{equation}
\vspace*{-0.2cm}

Similarly, we compute the mean distance scores of the $L-1$ previous frames for each row of diagonal entries in $D_\text{Diag}^j$, resulting in the mean distance scores vector $\mu_\text{Diag}^j$. We define the minimum sum rate $A_{1}^j$ as: 

\vspace*{-0.2cm}
\begin{equation} 
A_{1}^j = \frac{\sum_{\ell=1}^{L-1} D_\text{Diag}^j(d^*, \ell)}{\sum_{\ell=1}^{L-1} D_L^j(h^*, \ell) + \epsilon},
\end{equation}
where $d^*$ is the row of diagonal entries with the top $k$ lowest mean from $\mu_\text{Diag}^j$, $h^*$ is the row of horizontal entries with the top $k$ lowest mean from $\mu_\text{Hor}^j$ mean distance scores vector, and $\epsilon$ is added to prevent division by zero; alternatively, a max operation could be used.

\subsubsection{Minimum Value Rate}
To capture the reliability of the strongest match, we compute the ratio between the top $k$ lowest distance scores for the query and the distance score for the query at the row of diagonal entries with the lowest top $k$ sum:

\vspace*{-0.5cm}
\begin{equation} 
A_{2}^j = \frac{\text{sort}(D_{L}^j(:, L))[k]}{D_\text{Diag}^j(i^*, L) + \epsilon},
\end{equation}
where $i^*$ is the row of diagonal entries with the top $k$ lowest distance scores sum.

\subsubsection{Global Block Sum Rate}

To quantify the alignment stability of the top-ranked matches across the query's previous $L-1$ frames, we compute the ratio between the sum of the diagonal entries for the row with the top $k$ minimum diagonal sum and the sum of the diagonal block entries at the same row.
We define a block as a set of $L$ consecutive rows of diagonal entries, where each value in the block is replaced by the mean of the block. Specifically, a block is defined as:

\vspace*{-0.4cm}
\begin{equation}
    D_{\text{block},i}^{j} = \left\{ D^{j}_{\text{Diag}}(i + \ell, i + \ell) \;\middle|\; \ell = 0, \dots, L - 1 \right\}.
\end{equation}
\vspace*{-0.4cm}

Each element in the block is set to the block's mean, $\mu_\text{block, i}$:

\vspace*{-0.4cm}

\begin{equation}
    \quad D^{j}_{\text{Diag}}(i + \ell, i + \ell) = \mu^{j}_{\text{block},i}, \forall \ell \in \{0, \dots, L - 1\}.
\tag{7}
\end{equation}

Finally, we compute the global block sum rate $A_{3}^j$ as: 

\vspace*{-0.2cm}
\begin{equation}
    A_{3}^j = \frac{\sum_{\ell=1}^{L-1} D_\text{Diag}^j(i^*, \ell)}{\sum_{\ell=1}^{L} D_\text{block}^j(i^*, \ell)}.
\end{equation}

\subsubsection{Global Group Sum Rate}

Finally, we compute the ratio between the sum of the diagonal entries for the row with the top $k$ lowest diagonal sum, and the corresponding sum of the diagonal group entries in the same row. 
We define a diagonal group by averaging the values of the diagonal entries in a window centered around each row. Specifically, for each row, we consider the $W$ rows before and after it, and compute the mean of the values within this window. This process smoothens the diagonal by incorporating surrounding context.
We define a group as:

\vspace*{-0.2cm}
\begin{equation}
    D_\text{group}^j(i, :) = \frac{1}{2W+1} \sum_{m=\max(0, i-W)}^{\min(R, i+W+1)} D_\text{Diag}^j(m, :).
\end{equation}

Finally, we compute the global group sum rate as: 

\begin{equation}
    A_{4}^j = \frac{\sum_{\ell=1}^{L-1} D_\text{Diag}^j(i^*, \ell)}{\sum_{\ell=1}^{L} D_\text{group}^j(i^*, \ell)}.
\end{equation}

\subsubsection{Final Feature Representation}

We compute these attributes for each frame based on the top $K$ lowest entries, which are ranked by their sums. 
Hence, for each query frame, we extract four attributes for each top $k$ lowest entries. 
Note, we train and evaluate our MLP-based SMR predictor based only on the top $1$ attributes.

\subsection{SMR Predictor Supervised Learning Method}

For our SMR prediction model, we employ a supervised Multi-layer Perceptron (MLP) classifier. 
We construct the training dataset by using the query frame attributes from the training segment of each dataset, evaluated across all VPR models. 
For example, with $100$ query frames from Dataset $1$ and $7$ VPR models, this results in $100 \times 7 = 700$ training samples. The model is retrained separately on each dataset.

For each query frame, the predictor outputs a discrete label, $y \in \{0,1,2,3\}$, corresponding to the following outcomes for the query match:
\begin{itemize}
    \item $y=0$: Correct before and after sequence matching.
    \item $y=1$: Correct before and incorrect after sequence matching. 
    \item $y=2$: Incorrect before and correct after sequence matching. 
    \item $y=3$: Incorrect before and after sequence matching. 
\end{itemize}

\subsection{Match Removal}

Leveraging the predictor to improve performance involves removing initial proposed matches that are predicted as ``incorrect after sequence matching'' by our predictor ($y=1 \; \& \; y=3$). We note here that whilst ideally the prediction would be perfect, in the results section we show that even an imperfect predictor is able to produce a significant overall improvement in VPR performance. By using our predictor to remove some matches, the aim is to primarily remove actual false positives, whilst minimizing the number of true positives that are removed (which turns them into false negatives). 

For our implementation, we apply the predictor at the last operating point threshold of the underlying VPR + sequence matching system, where the recall of the VPR system with sequence matching is maximal. While the experimental results will shed more light on the interactions of the predictor with the underlying system, at an intuitive level, it can only (potentially) increase precision and reduce recall. So, starting at the maximal recall point is an appropriate starting point.

\subsection{Match Restoration}

If matches can be removed because they are likely wrong, %
can we promote other highly ranked matches to replace them? That is the aim of this investigation, where we use our predictor on the next top $K$ ranked matches (as ranked by the distance scores of the VPR system), and replace the ``removed'' false positive with the top ranked match that has the highest score from the predictor system. The aim of such an approach is to provide a mechanism for increasing recall, whilst minimizing the drop in precision -- the inverse of our main contribution. We provide this analysis here not as part of our main methodological contribution, which already by itself improves VPR performance across many methods and datasets without this latter step -- but rather to provide some initial insights into this natural follow-on question.

\section{Experimental Setup}

\subsection{Implementation Details}

We normalize the distance score matrix, $D_L^j$, by dividing each entry by the maximum absolute value in $D_L^j$.
As the number of query samples per output class varies, we used the Synthetic Minority Over-sampling Technique (SMOTE)~\cite{chawla2002smote} to upsample minority classes for training only. 
We use $W=2$ for the Global Group Sum Ratio attribute.
Our MLP classifier has three hidden layers of 128 neurons each, with the Rectified Linear Units (ReLU) activation and Adam optimization. The initial learning rate is set to $0.0001$, and $\alpha=0.001$. We apply Stratified K-Fold cross-validation and evaluate using the macro F1-score. 
We implemented our classifier using the SkLearn Python library~\cite{scikit-learn} on a machine with an i7 CPU, RTX 3080 GPU.

\subsection{VPR Techniques}

We selected several VPR techniques to show the applicability of our approach, which are CosPlace~\cite{berton2022rethinking}, EigenPlaces~\cite{berton2023eigenplaces}, MixVPR~\cite{ali2023mixvpr}, SALAD~\cite{izquierdo2024optimal}, NetVLAD~\cite{Arandjelovic2018}, AP-GeM~\cite{revaud2019learning}, and Sum-of-Absolute Differences (SAD)~\cite{milford2012seqslam}. 

CosPlace~\cite{berton2022rethinking} uses a classification-based training strategy, iterates over different non-overlapping groups of places, and outputs global descriptors at inference. 
EigenPlaces~\cite{berton2023eigenplaces} trains using multiple different viewpoints of the same place to create robust feature descriptors. 
MixVPR~\cite{ali2023mixvpr} processes features extracted from a CNN-based architecture through a cascade of Feature Mixer layers. 
SALAD~\cite{izquierdo2024optimal} frames the feature-to-cluster assignment as an optimal transport problem and uses a fine-tuned foundation model, DINOv2.
NetVLAD~\cite{Arandjelovic2018} aggregates local features into global descriptors using VLAD with learnable weights to generate global descriptors. 
Sum-of-Absolute Differences (SAD)~\cite{milford2012seqslam} is a non-learned model using per-pixel differences as the feature descriptors.

\subsection{Datasets}

We used the following datasets:

\textbf{The Nordland dataset~\cite{sunderhauf2013we}:} This 728 km train route in Norway spans across four seasons: spring, summer, fall, and winter. Segments with speeds below 15 km/h were excluded as done in~\cite{camara2020visual}. 
We used the summer traverse as reference and winter as query.
We used $500$ reference and corresponding query images for training, $500$ geographically separate images for validation, and $9000$ for testing.

\textbf{The Oxford RobotCar dataset~\cite{RobotCar}:} 
This dataset includes over 100 traverses in Oxford under various conditions, such as different times of day and seasons. 
We used the front left stereo frames from the Rain traverse, 2015-10-29-12-18-17, as reference and from the Dusk traverse, 2014-11-21-16-07-03, as query as in~\cite{molloy2020intelligent}.
We used $500$ reference and query images for training, $500$ geographically separate images for validation, and $1000$ for testing.

\textbf{SFU Mountain dataset~\cite{bruce2015sfu}:}
This dataset with 385 frames was recorded by a mobile robot in a semi-structured bushland terrain at Burnaby Mountain, British Columbia, Canada, under challenging appearance conditions. 
As in~\cite{neubert2021hyperdimensional}, we used the right stereo frames and used the dry traverse as reference and the dusk traverse as query.
We used the first $100$ reference and query images for training, $100$ geographically separate images for validation, and $185$ for testing.

\subsection{Performance Metrics}

We assess the performance of our SMR predictor when integrated into a VPR system with sequence matching, to remove matches predicted as incorrect after sequence matching (first filter), and then to replace the removed matches with the top $K$ ranked matches (second filter).

We use precision, recall, and F1-score performance metrics, derived from true positives (correct matches), false positives (incorrect matches), and false negatives (missed matches).
We also use the area under the precision-recall curve (PR AUC), and the area over the precision-recall curve (PR AOC) as they are succinct statistical summaries of the performance. We are particularly interested in the change in the area \textbf{over} the curve, as it directly represents the error rate across the full operating range (from low to high recall) of the systems being analyzed (and consequently, reductions in this area represent reductions in the error rate).

For all three datasets, we use a ground truth tolerance of $\pm 2$ frames.
We report the area under the precision-recall curve using the maximum recall that our predictor, for match removal, obtains for a particular VPR model evaluated on a particular dataset. 
We note here that whilst different bodies of research have often used slightly varying parameters by which to evaluate their systems, the primary measurement of system effectiveness here is the \textit{relative} change in performance of the proposed method, compared to the baseline systems, regardless of the exact specific parameters like ground-truth tolerance used.

\section{results}

This section demonstrates the performance of our sequence matching receptiveness predictor, both independently and in terms of its impact on VPR performance when used for match removal.

\begin{table*}[t]
\vspace*{0.2cm}
\caption{Area Over Precision Recall curve comparison at max recall for each VPR method on Nordland, Oxford RobotCar, and SFU Mountain datasets (test set of each dataset).}
\label{tab:main_table}
\centering
\begin{tabular}{l|c|c|c|c|c|c|c|c|c|c}
\multicolumn{2}{c|}{} & AP-GeM & CosPlace & EigenPlaces & MixVPR & NetVLAD & SAD & SALAD & Mean & Improved\\
\hline
               & VPR+SM        & 0.6522 & 0.0220 & 0.0155 & 0.0061 & 0.2478 & 0.2164 & 0.0017 & 0.1660 &  \\ 
Nordland & VPR+SM+Pred   & 0.4862 & 0.0136 & 0.0106 & 0.0042 & 0.1274 & 0.1050 & 0.0011 & 0.1069 & \\ 
               & Reduction (\%)    & \textcolor{green}{25.46} & \textcolor{green}{38.17} & \textcolor{green}{31.31} & \textcolor{green}{31.96} & \textcolor{green}{48.61} & \textcolor{green}{51.46} & \textcolor{green}{37.55} & \textcolor{darkgreen}{37.79} & 7 / 7 \\ 
\hline
               & VPR+SM        & 0.0950 & 0.0842 & 0.0867 & 0.0363 & 0.0519 & 0.1479 & 0.0769 & 0.0827 &  \\ 
Oxford RobotCar & VPR+SM+Pred   & 0.1085 & 0.0880 & 0.0682 & 0.0257 & 0.0312 & 0.0975 & 0.0561 & 0.0679 & \\ 
               & Reduction (\%)    & \textcolor{red}{-14.22} & \textcolor{red}{-4.45} & \textcolor{green}{21.31} & \textcolor{green}{29.34} & \textcolor{green}{39.83} & \textcolor{green}{34.07} & \textcolor{green}{26.95} & \textcolor{darkgreen}{18.97} & 5 / 7 \\ 
\hline
               & VPR+SM        & 0.0455 & 0.0000 & 0.0000 & 0.0000 & 0.0084 & 0.0108 & 0.0000 & 0.0092 &  \\ 
SFU Mountain & VPR+SM+Pred   & 0.0331 & 0.0000 & 0.0000 & 0.0000 & 0.0025 & 0.0049 & 0.0000 & 0.0058 & \\ 
               & Reduction (\%)    & \textcolor{green}{27.18} & \textcolor{black}{0.00} & \textcolor{black}{0.00} & \textcolor{black}{0.00} & \textcolor{green}{70.17} & \textcolor{green}{54.40} & 0.00 & \textcolor{darkgreen}{21.68} & 3 / 7 \\ 
\hline

\end{tabular}
\end{table*}

\subsection{Performance of our SMR Predictor}
\label{smr_predictor}

This section presents the results for the scenario, where our SMR predictor is used to remove matches predicted as ``incorrect after sequence matching'' in a VPR system with sequence matching using a sequence length of 4 frames.

\Cref{tab:main_table} presents the area over the precision-recall curves (AOC) for VPR models with sequence matching, both with and without our predictor. Specifically, it shows the AOC of the VPR models when sequence matching is applied alone and when our predictor is incorporated to remove matches predicted as incorrect after sequence matching. Additionally, we report the percentage reduction in AOC when incorporating our predictor compared to the baseline VPR system with sequence matching.

The overall average reduction in AOC across all VPR methods and datasets, when our predictor is incorporated into the VPR systems with sequence matching, is $26.15\%$. %

In 15 out of 21 evaluated VPR method and dataset configurations, using our predictor results in an AOC reduction compared to the VPR system with sequence matching. 
In 4 out of 21 cases, all from the SFU Mountain dataset, the AUC-PR of the VPR system with sequence matching is already capped at 100\%. 
We show that applying our predictor maintains this perfect performance, ensuring no degradation in the VPR+SM system's performance.
In 2 out of 21 cases, both from the Oxford RobotCar dataset, there is a slight performance degradation, with the AOC increasing from $0.095$ to $0.11$ for AP-GeM and from $0.084$ to $0.088$ for CosPlace.

The macro F1-score of our SMR predictor in isolation across all seven VPR models on the test queries was $0.54$ on Nordland, $0.37$ on Oxford RobotCar, and $0.49$ on the SFU-Mountain dataset.
While our predictor is not perfect (and perfect prediction, of course, is unlikely to be achievable), its current level of performance is good enough to significantly improve the average performance of the VPR system after match removal.~\Cref{discussion} details potential improvements to further enhance its performance, which would likely then generate even more improvements in the VPR performance.

\subsection{Ablation Study: Varying Sequence Length}
\label{diff_se}

\begin{figure*}[htpb]
  \centering
  \vspace*{-0.2cm}
  
  \hfill
  \includegraphics[width=0.45\linewidth,trim={5mm 4mm 0mm 2mm},clip]{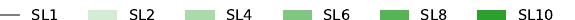} 
  \hfill
  \includegraphics[width=0.45\linewidth,trim={5mm 4mm 0mm 2mm},clip]{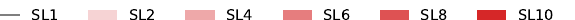} \\
  \vspace{-0.3cm}

  \subfloat[First Filter\label{fig:first_filter}]{\includegraphics[width=0.48\linewidth,trim={1mm 3mm 2mm 2mm},clip]{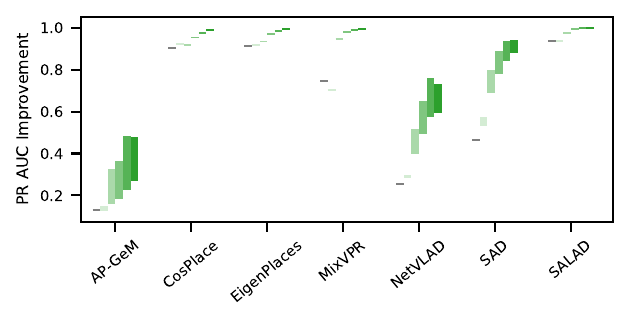}} \hfill
  \subfloat[Second Filter\label{fig:second_filter}]{\includegraphics[width=0.48\linewidth,trim={1mm 3mm 0mm 2mm},clip]{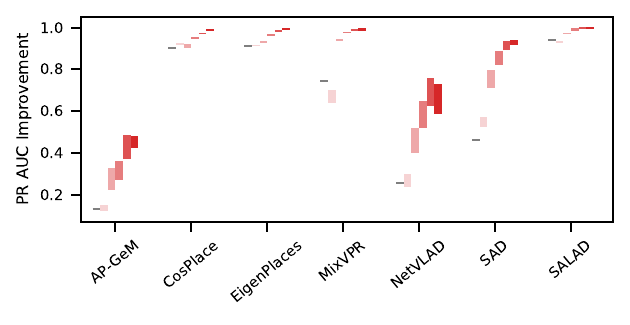}}

  \caption{\textbf{(Left)} Ablation study on the impact of sequence length on PR AUC improvement when applying the first filter (VPR+SM to VPR+SM+pred), showing that AUC increases (green bars). \textbf{(Right) }Effect of the second filter, which replaces the removed incorrect matches with predicted highly-ranked matches. The PR AUC decreases (red bars) compared to the first filter at maximum recall.}
  \vspace*{-0.3cm}
  \label{fig:pred_f1_pr}
\end{figure*}

We analyze the influence of sequence length in sequence matching on our SMR predictor. 
\Cref{fig:first_filter} shows the results for all seven VPR models evaluated on the Nordland dataset. 
As in~\Cref{smr_predictor}, we use our predictor to remove matches predicted as incorrect after sequence matching in the VPR system with sequence matching. 
The bottom of each bar represents the PR AUC of the VPR system with sequence matching, while the top indicates the PR AUC when our predictor is applied to remove incorrect matches. If the difference is positive, the bar is green (as in all results in \Cref{fig:first_filter}); otherwise, it would be red. 

We trained our SMR predictor separately for sequence lengths of 2, 4, 6, 8, and 10, which define how many prior query frames are considered when generating the attributes for our predictor.
The results show that our predictor consistently maintains or improves the PR AUC of the VPR system with sequence matching across different sequence lengths: demonstrating that the method is useful across a wide range of operating modes (sequence matching lengths) that a user may choose for their specific application.

\subsection{Ablation Study: Varying Predictor's Trust (First Filter)}

\begin{figure}[t]
  \centering

  \hfill
  \includegraphics[width=0.49\linewidth,trim={5mm 4mm 0mm 4mm},clip]{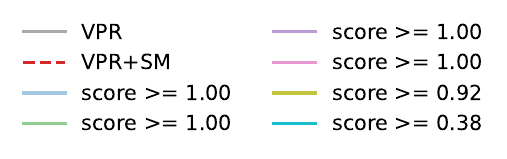} 
  \hfill
  \includegraphics[width=0.49\linewidth,trim={5mm 4mm 4mm 4mm},clip]{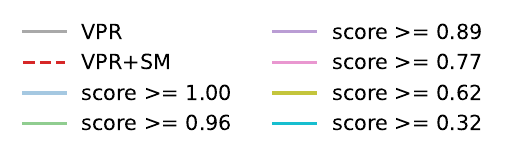} \\

  \vspace*{-0.25cm}
  \subfloat[EigenPlaces]{\includegraphics[width=0.48\linewidth,trim={1mm 3mm 2mm 2mm},clip]{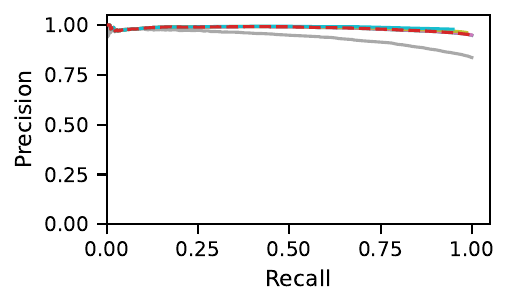}}%
  \hfill
  \subfloat[NetVLAD]{\includegraphics[width=0.48\linewidth,trim={2mm 3mm 2mm 2mm},clip]{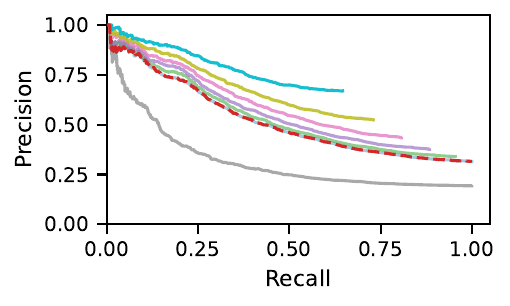}}

  \caption{Ablation Study: Varying predictor's trust (VPR+SM+Pred -- first filter) using EigenPlaces and NetVLAD on Nordland dataset.}
  \vspace*{-0.5cm}
  \label{fig:abl_pred_trust}
\end{figure}

We evaluate how varying the trust level in our SMR predictor affects the precision-recall plot of the VPR system with sequence matching (4-frame sequence length).
\Cref{fig:abl_pred_trust} shows the PR plot of our SMR predictor at different trust levels for match removal on EigenPlaces and NetVLAD. 
When we only trust the predictor at full confidence ($\text{score}\ge 1$), very few to no matches are removed. In this case, the effect of our predictor is minimal, and the PR plot closely mirrors that of the VPR system with sequence matching.

For NetVLAD, lowering the trust threshold allows the predictor to remove more incorrect matches after sequence matching, resulting in a clear gain in precision at lower recall values. 
In contrast, for EigenPlaces, the benefit is marginal across trust levels, with the precision-recall curves closely placed.
This suggests that in a high-performing VPR model (like EigenPlaces) with sequence matching, it is more challenging to remove the few remaining incorrect matches, as they are already sparse, while effectively retaining the correct matches.
As the predictor removes matches predicted as incorrect after sequence matching, the resulting PR curve does not extend to a recall of $100\%$.
\Cref{res:match_res} examines how our predictor performs when it is first used to remove incorrect matches after sequence matching and then used as a second filter to replace the removed match with another highly ranked match.

\subsection{Ablation Study: Impact of Each Attribute}

In~\Cref{tab:abl_mlp_att}, we performed an ablation study on the Nordland dataset to assess the impact of each attribute individually on the F1-score of our MLP-based SMR predictor in isolation, i.e. without applying it to the VPR system with sequence matching. 
Results are averaged across all seven VPR models on the Nordland dataset.
Using only A1 or A2 resulted in an F1-score of $0.42$, while using A3 and A4 resulted in F1-scores of $0.40$ and $0.37$, respectively.

Using all four attributes improved performance to $0.54$.
This demonstrates that incorporating all four attributes improves classifier performance, validating our approach.
These attributes were selected through an iterative process of visualizing distance scores, analyzing top-K matches, deriving descriptive statistics, and examining class-wise sample distributions.

\begin{table}[t]
\vspace*{0.2cm}
\caption{Ablation Study: Contributions of the MLP attributes based on the predictor's F1-score in isolation, averaged across all VPR methods on Nordland dataset (Test set).}
\label{tab:abl_mlp_att}
\centering
\begin{tabular}{c|c|c|c|c}
A1 & A2 & A3 & A4 & F1-score \\
\hline
$\checkmark$ & $\times$ & $\times$ & $\times$ & 0.42 \\
$\times$ & $\checkmark$ & $\times$ & $\times$ & 0.42 \\
$\times$ & $\times$ & $\checkmark$ & $\times$ & 0.40 \\
$\times$ & $\times$ & $\times$ & $\checkmark$ & 0.37 \\
$\checkmark$ & $\checkmark$ & $\checkmark$ & $\checkmark$ & 0.54 \\

\end{tabular}
\end{table}

\subsection{Qualitative Results}

\begin{figure}[t]
  \centering
  \subfloat{\includegraphics[width=0.99\linewidth,trim={0mm 1mm 1mm 1mm},clip]{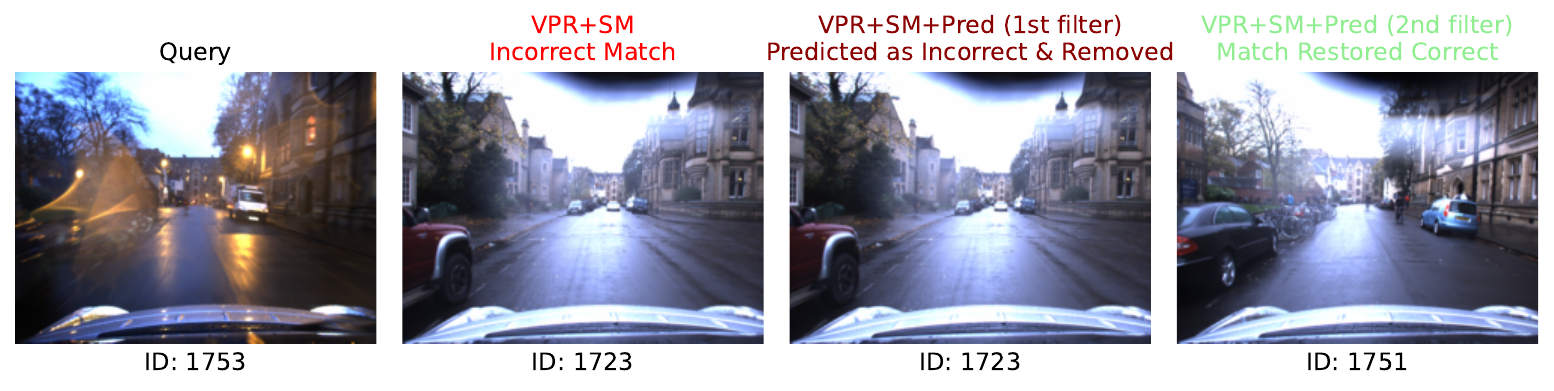}}\\
  \vspace{-0.35cm}
  \subfloat{\includegraphics[width=0.99\linewidth,trim={0mm 1mm 1mm 1mm},clip]{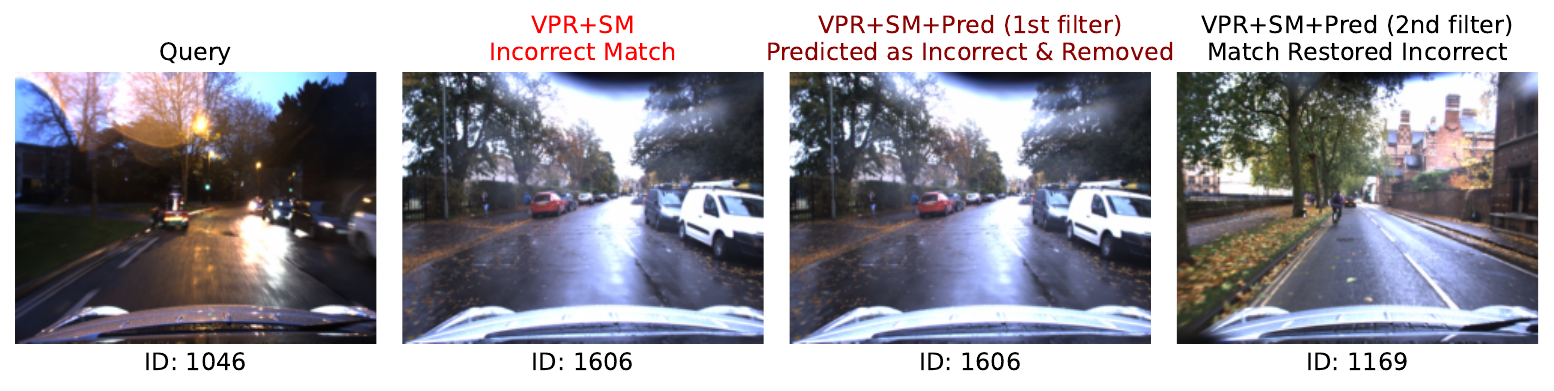}}\\
  \vspace{-0.35cm}
  \subfloat{\includegraphics[width=0.99\linewidth,trim={0mm 1mm 1mm 1mm},clip]{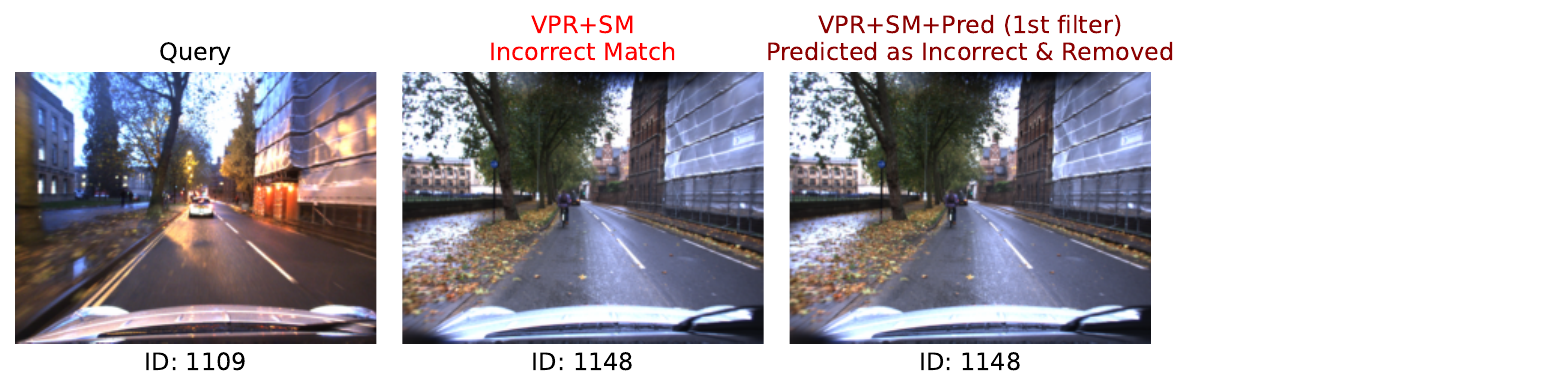}}\\
  \vspace{-0.35cm}
  \subfloat{\includegraphics[width=0.99\linewidth,trim={0mm 1mm 1mm 1mm},clip]{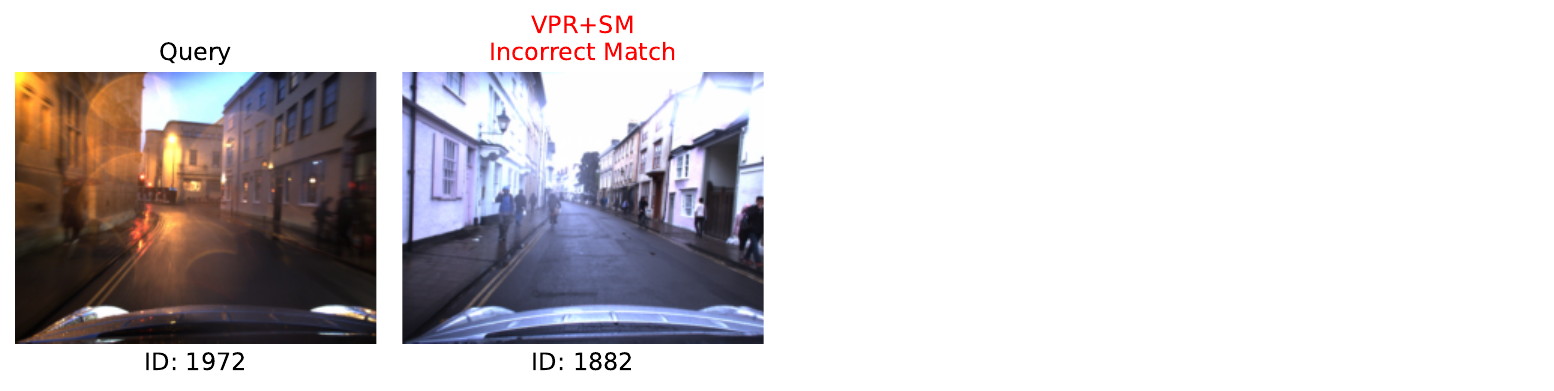}}\\

  \caption{
  Qualitative examples of our predictor applied to the baseline VPR system with sequence matching (NetVLAD on Oxford RobotCar, sequence length 4) to remove and restore matches predicted as incorrect. Cases of successful corrections and failures are shown.
  }
  \vspace*{-0.5cm}

  \label{fig:qualitative}
\end{figure}

We present qualitative results in \Cref{fig:qualitative}, showing examples where the baseline VPR system with sequence matching produces incorrect predictions. 
We show cases where our predictor effectively removes incorrect matches and restores them with another highly-ranked match, and instances where the predictor fails to detect or restore incorrect matches.

\subsection{Match Restoration}
\label{res:match_res}

Can discarded matches be replaced by alternative correct matches? 
This section presents the results for applying our predictor a second time, this time to rank the alternative highly ranked matches for a discarded match to replace it with the most credible alternative among the next top three ranked matches.

In \Cref{fig:first_filter}, our predictor removes matches predicted as incorrect after sequence matching (match removal) on the Nordland dataset for all seven VPR models. \Cref{fig:second_filter} shows results on the same dataset when our predictor is also used to replace incorrect matches with the highest-confidence top-ranked match (match restoration).
The bars represent the PR AUC difference between using our predictor for match removal versus match restoration. 
A positive difference (green) would indicate an improvement, but all results here are negative, as shown in red, showing that match restoration lowers PR AUC compared to that of match removal.

Restoring matches appears more challenging than removing incorrect ones, as it requires higher predictor performance for a significant positive impact. 
Nevertheless, our predictor’s PR AUC in match restoration is similar to the baseline VPR system with sequence matching, with mean PR AUC values across all VPR models of $0.72$ (ours) and $0.71$ (baseline) on Nordland, $0.55$ (both) on Oxford RobotCar, and $0.89$ (ours) and $0.92$ (baseline) on SFU Mountain.
We also investigated varying the match restoration threshold, which showed that higher thresholds restore fewer but more confident matches, while lower thresholds restore more matches. The majority of replaced matches fall between thresholds 0.91 and 1.0.

\subsection{Inference Compute Time}
\label{res:inf_time}

The average inference time of our predictor, considering a sequence length of 4 frames, is $0.54$ ms, with the majority of the inference time $>0.5$ ms spent on computing the attributes of the predictor, noting our code has not been optimized.

\section{Discussion}
\label{discussion}

We presented a novel approach for predicting the receptiveness of single-frame VPR methods to distance-based sequence matching, which is independent of the underlying VPR model. By removing matches output by the sequence matching system that are predicted as incorrect, we can substantially improve the average VPR performance of the underlying VPR + Sequence Matching system, a result demonstrated across seven VPR models and three benchmark datasets. Importantly, the resulting performance improvement is superior to simply choosing a different operating point on the baseline system's precision-recall curve -- that is, performance after applying the predictor moves above the baseline precision-recall curve in most cases. While the predictor mechanism used in this paper, an MLP, is relatively straightforward, it is particularly interesting to note that even with imperfect prediction performance, such a substantial average improvement in VPR performance can be achieved.

We believe this work is a promising expansion on the recent single-frame introspective work (covered in~\Cref{LR:introspect}) to introspection on widely used filtering and sequence-based methods. Intuitively, high-performance introspection may be more achievable on sequential information over time, as there is more information available, and in robotics applications, there are interesting correlations across sequences in this temporal data. Some areas for future exploration would be investigating the use of modern machine learning techniques with particular mechanistic analogues to this process, such as Transformers.

While in this work significant average VPR performance improvement was achievable by applying the predictor to remove some  matches that are predicted to be incorrect, there is great appeal in potentially \textit{restoring} lower ranked candidate matches. We investigated this possibility in this paper, showing that match restoration appears to be a fundamentally more difficult task, at least in the context of providing \textit{further} performance improvements over the main predictor-filter approach demonstrated in this work. More sophisticated predictor learning schemes may be able to obtain the predictor performance required to achieve further VPR performance improvements: in parallel simulations (not presented in this paper) we have discovered that the predictor performance required to beneficially restore matches is much higher than to filter out bad matches.

Computational efficiency was not the primary aim of the research presented here, although~\Cref{res:inf_time} presents a brief description of the inference computation that shows this is a very lightweight process. 
The current system is a post-sequence-matching process, in that it runs after sequence matching has been performed.
Future work could investigate the possibility of using our predictor in an incremental manner (e.g. using shorter sequences and only on a subset of the data) to guide sequence matching before it is fully computed, enabling both potential computational benefits and matching latency reductions, with the predictor being applied live, e.g. to determine whether increasing the sequence length at that moment in time is worthwhile.

\bibliographystyle{IEEEtran}
\bibliography{references}

\begin{thebibliography}{10}
\providecommand{\url}[1]{#1}
\csname url@rmstyle\endcsname
\providecommand{\newblock}{\relax}
\providecommand{\bibinfo}[2]{#2}
\providecommand\BIBentrySTDinterwordspacing{\spaceskip=0pt\relax}
\providecommand\BIBentryALTinterwordstretchfactor{4}
\providecommand\BIBentryALTinterwordspacing{\spaceskip=\fontdimen2\font plus
\BIBentryALTinterwordstretchfactor\fontdimen3\font minus
  \fontdimen4\font\relax}
\providecommand\BIBforeignlanguage[2]{{%
\expandafter\ifx\csname l@#1\endcsname\relax
\typeout{** WARNING: IEEEtran.bst: No hyphenation pattern has been}%
\typeout{** loaded for the language `#1'. Using the pattern for}%
\typeout{** the default language instead.}%
\else
\language=\csname l@#1\endcsname
\fi
#2}}

\bibitem{tsintotas2022revisiting}
K.~A. Tsintotas, L.~Bampis, and A.~Gasteratos, ``The revisiting problem in
  simultaneous localization and mapping: A survey on visual loop closure
  detection,'' \emph{Trans. Intell. Transp. Syst.}, vol.~23, no.~11, pp.
  19\,929--19\,953, 2022.

\bibitem{lajoie2022towards}
P.-Y. Lajoie \emph{et~al.}, ``Towards collaborative simultaneous localization
  and mapping: a survey of the current research landscape,'' \emph{Field
  Robot.}, vol.~2, pp. 971--1000, 2022.

\bibitem{yan2024gs}
C.~Yan \emph{et~al.}, ``Gs-slam: Dense visual slam with 3d gaussian
  splatting,'' in \emph{IEEE Conf. Comput. Vis. Pattern Recog.}, 2024, pp.
  19\,595--19\,604.

\bibitem{zhu2024sni}
S.~Zhu \emph{et~al.}, ``Sni-slam: Semantic neural implicit slam,'' in
  \emph{IEEE Conf. Comput. Vis. Pattern Recog.}, 2024, pp. 21\,167--21\,177.

\bibitem{lecun2015deep}
Y.~LeCun, Y.~Bengio, and G.~Hinton, ``Deep learning,'' \emph{Nat.}, vol. 521,
  no. 7553, pp. 436--444, 2015.

\bibitem{han2022survey}
K.~Han \emph{et~al.}, ``A survey on vision transformer,'' \emph{IEEE Trans.
  Pattern Anal. Mach. Intell.}, vol.~45, no.~1, pp. 87--110, 2022.

\bibitem{vaswani2017attention}
A.~Vaswani \emph{et~al.}, ``Attention is all you need,'' \emph{Adv. Neural
  Inform. Process. Syst.}, vol.~30, 2017.

\bibitem{alayrac2022flamingo}
J.-B. Alayrac \emph{et~al.}, ``Flamingo: a visual language model for few-shot
  learning,'' \emph{Adv. Neural Inform. Process. Syst.}, vol.~35, pp.
  23\,716--23\,736, 2022.

\bibitem{masone2021survey}
C.~Masone and B.~Caputo, ``A survey on deep visual place recognition,''
  \emph{IEEE Access}, vol.~9, pp. 19\,516--19\,547, 2021.

\bibitem{zhang2021visual}
X.~Zhang, L.~Wang, and Y.~Su, ``Visual place recognition: A survey from deep
  learning perspective,'' \emph{Pattern Recognit.}, vol. 113, p. 107760, 2021.

\bibitem{milford2012seqslam}
M.~J. Milford and G.~F. Wyeth, ``{SeqSLAM}: Visual route-based navigation for
  sunny summer days and stormy winter nights,'' in \emph{IEEE Int. Conf. Robot.
  Autom.}, 2012, pp. 1643--1649.

\bibitem{schubert2021fast}
S.~Schubert, P.~Neubert, and P.~Protzel, ``Fast and memory efficient graph
  optimization via icm for visual place recognition.'' in \emph{Robot. Sci.
  Syst.}, vol.~73, 2021.

\bibitem{naseer2014robust}
T.~Naseer \emph{et~al.}, ``Robust visual robot localization across seasons
  using network flows,'' in \emph{AAAI Conf. Artif. Intell.}, vol.~28, no.~1,
  2014.

\bibitem{garg2022seqmatchnet}
S.~Garg, M.~Vankadari, and M.~Milford, ``{SeqMatchNet}: Contrastive learning
  with sequence matching for place recognition \& relocalization,'' in
  \emph{Conference on Robot Learning}, 2022, pp. 429--443.

\bibitem{mereu2022learning}
R.~Mereu \emph{et~al.}, ``Learning sequential descriptors for sequence-based
  visual place recognition,'' \emph{IEEE Robot. Autom. Lett.}, vol.~7, no.~4,
  pp. 10\,383--10\,390, 2022.

\bibitem{garg2021seqnet}
S.~Garg and M.~Milford, ``{SeqNet}: Learning descriptors for sequence-based
  hierarchical place recognition,'' \emph{IEEE Robot. Autom. Lett.}, vol.~6,
  no.~3, pp. 4305--4312, 2021.

\bibitem{arroyo2015towards}
R.~Arroyo \emph{et~al.}, ``Towards life-long visual localization using an
  efficient matching of binary sequences from images,'' in \emph{IEEE Int.
  Conf. Robot. Autom.}, 2015, pp. 6328--6335.

\bibitem{schubert2023visual}
S.~Schubert \emph{et~al.}, ``Visual place recognition: A tutorial,'' \emph{IEEE
  Robot. Autom. Mag.}, 2023.

\bibitem{carson2022predicting}
H.~Carson, J.~J. Ford, and M.~Milford, ``Predicting to improve: Integrity
  measures for assessing visual localization performance,'' \emph{IEEE Robot.
  Autom. Lett.}, vol.~7, no.~4, pp. 9627--9634, 2022.

\bibitem{claxton2024improving}
O.~Claxton \emph{et~al.}, ``Improving visual place recognition based robot
  navigation by verifying localization estimates,'' \emph{IEEE Robot. Autom.
  Lett.}, 2024.

\bibitem{carson2023unsupervised}
H.~Carson, J.~J. Ford, and M.~Milford, ``Unsupervised quality prediction for
  improved single-frame and weighted sequential visual place recognition,'' in
  \emph{IEEE Int. Conf. Robot. Autom.}, 2023, pp. 3955--3961.

\bibitem{jegou2010aggregating}
H.~J{\'e}gou \emph{et~al.}, ``Aggregating local descriptors into a compact
  image representation,'' in \emph{IEEE Conf. Comput. Vis. Pattern Recog.},
  2010, pp. 3304--3311.

\bibitem{Arandjelovic2018}
R.~Arandjelovic \emph{et~al.}, ``{NetVLAD: CNN} architecture for weakly
  supervised place recognition,'' \emph{IEEE Trans. Pattern Anal. Mach.
  Intell.}, vol.~40, no.~6, pp. 1437--1451, 2018.

\bibitem{berton2022rethinking}
G.~Berton, C.~Masone, and B.~Caputo, ``Rethinking visual geo-localization for
  large-scale applications,'' in \emph{IEEE Conf. Comput. Vis. Pattern Recog.},
  2022, pp. 4878--4888.

\bibitem{berton2023eigenplaces}
G.~Berton \emph{et~al.}, ``Eigenplaces: Training viewpoint robust models for
  visual place recognition,'' in \emph{IEEE Conf. Comput. Vis. Pattern Recog.},
  2023, pp. 11\,080--11\,090.

\bibitem{ali2023mixvpr}
A.~Ali-Bey, B.~Chaib-Draa, and P.~Giguere, ``Mixvpr: Feature mixing for visual
  place recognition,'' in \emph{IEEE/CVF Winter Conf. Appl. Comput. Vis.},
  2023, pp. 2998--3007.

\bibitem{izquierdo2024optimal}
S.~Izquierdo and J.~Civera, ``Optimal transport aggregation for visual place
  recognition,'' in \emph{IEEE Conf. Comput. Vis. Pattern Recog.}, 2024, pp.
  17\,658--17\,668.

\bibitem{lu2024cricavpr}
F.~Lu \emph{et~al.}, ``Cricavpr: Cross-image correlation-aware representation
  learning for visual place recognition,'' in \emph{IEEE Conf. Comput. Vis.
  Pattern Recog.}, 2024, pp. 16\,772--16\,782.

\bibitem{keetha2023anyloc}
N.~Keetha \emph{et~al.}, ``Anyloc: Towards universal visual place
  recognition,'' \emph{IEEE Robot. Autom. Lett.}, vol.~9, no.~2, pp.
  1286--1293, 2023.

\bibitem{al2020localization}
J.~Al~Hage \emph{et~al.}, ``Localization integrity for intelligent vehicles
  through fault detection and position error characterization,'' \emph{Trans.
  Intell. Transp. Sys.}, vol.~23, no.~4, pp. 2978--2990, 2020.

\bibitem{zhu2022integrity}
C.~Zhu, M.~Meurer, and C.~G{\"u}nther, ``Integrity of visual
  navigation—developments, challenges, and prospects,'' \emph{J. Inst.
  Navig.}, vol.~69, no.~2, 2022.

\bibitem{gautam2022method}
A.~Gautam \emph{et~al.}, ``A method for designing autonomous robots that know
  their limits,'' in \emph{IEEE Int. Conf. Robot. Autom.}, 2022, pp. 121--127.

\bibitem{arana2020localization}
G.~D. Arana \emph{et~al.}, ``Localization safety validation for autonomous
  robots,'' in \emph{IEEE/RSJ Int. Conf. Intell. Robot. Syst.}, 2020, pp.
  6276--6281.

\bibitem{guruau2018learn}
C.~Gur{\u{a}}u \emph{et~al.}, ``Learn from experience: Probabilistic prediction
  of perception performance to avoid failure,'' \emph{Int. J. Robot. Res.},
  vol.~37, no.~9, pp. 981--995, 2018.

\bibitem{chawla2002smote}
N.~V. Chawla \emph{et~al.}, ``Smote: synthetic minority over-sampling
  technique,'' \emph{J. Art. Intell. Res.}, vol.~16, pp. 321--357, 2002.

\bibitem{scikit-learn}
F.~Pedregosa \emph{et~al.}, ``Scikit-learn: Machine learning in {P}ython,''
  \emph{Journal of Machine Learning Research}, vol.~12, pp. 2825--2830, 2011.

\bibitem{revaud2019learning}
J.~Revaud \emph{et~al.}, ``Learning with average precision: Training image
  retrieval with a listwise loss,'' in \emph{Int. Conf. Comput. Vis.}, 2019,
  pp. 5107--5116.

\bibitem{sunderhauf2013we}
N.~S{\"u}nderhauf, P.~Neubert, and P.~Protzel, ``Are we there yet? {Challenging
  SeqSLAM} on a 3000 km journey across all four seasons,'' in \emph{IEEE Int.
  Conf. Robot. Autom. Worksh.}, 2013.

\bibitem{camara2020visual}
L.~G. Camara and L.~P{\v{r}}eu{\v{c}}il, ``Visual place recognition by spatial
  matching of high-level {CNN} features,'' \emph{Rob. Auton. Syst.}, vol. 133,
  p. 103625, 2020.

\bibitem{RobotCar}
W.~Maddern \emph{et~al.}, ``1 year, 1000 km: The {Oxford RobotCar} dataset,''
  \emph{Int. J. Robot. Res.}, vol.~36, no.~1, pp. 3--15, 2017.

\bibitem{molloy2020intelligent}
T.~L. Molloy \emph{et~al.}, ``Intelligent reference curation for visual place
  recognition via bayesian selective fusion,'' \emph{IEEE Robot. Autom. Lett.},
  vol.~6, no.~2, pp. 588--595, 2020.

\bibitem{bruce2015sfu}
J.~Bruce, J.~Wawerla, and R.~Vaughan, ``The {SFU} mountain dataset:
  Semi-structured woodland trails under changing environmental conditions,'' in
  \emph{IEEE Int. Conf. Robot. Autom.}, 2015.

\bibitem{neubert2021hyperdimensional}
P.~Neubert and S.~Schubert, ``Hyperdimensional computing as a framework for
  systematic aggregation of image descriptors,'' in \emph{IEEE Conf. Comput.
  Vis. Pattern Recog.}, 2021, pp. 16\,938--16\,947.

\end{thebibliography}

\end{document}